\documentclass{gretsi}

\usepackage[latin1]{inputenc}
\usepackage[french,english]{babel}
\usepackage{graphicx,afterpage}
\usepackage{fancybox}
\usepackage{psboxit,pstricks,rotating}
\usepackage{psfrag}
\usepackage{ulem}
\usepackage{amsmath,bbm} 
\usepackage{amsfonts} 
\usepackage{amssymb}
\usepackage{vector}
\usepackage{floatflt}
\usepackage{theorem}
\usepackage{stmaryrd}
\usepackage{euscript}
\usepackage{pifont}
\usepackage{subfigure}
\usepackage{array}
\usepackage[T1]{fontenc}
\usepackage{endnotes}

\title{G\'en\'eration de base de donn\'ees images IR sous contraintes avec variabilit\'e thermique intrins\`eque des cibles}
\auteur{\coord{J\'er\^ome}{Gilles}{1}, \coord{St\'ephane}{Landeau}{2}, \coord{Tristan}{Dagobert}{2}, \coord{Philippe}{Chevalier}{3}, \coord{Christian}{Bolut}{3}}
\adresse{\affil{1}{DGA-CEP/EORD, 16bis rue Prieur de la C\^ote d'Or 94110 Arcueil}\affil{2}{DGA-CEP/CGN, 16bis rue Prieur de la C\^ote d'Or 94110 Arcueil}\affil{3}{DGA-ETAS, BP 60036, Montreuil-Juign\'e 49245 Avrill\'e Cedex}}
\email{jerome.gilles@etca.fr, stephane.landeau@dga.defense.gouv.fr,\\tristan.dagobert@dga.defense.gouv.fr, philippe.chevalier@dga.defense.gouv.fr}

\frenchabstract{Dans cette communication, nous proposons une m\'ethode permettant de simuler des images de cibles en imagerie infrarouge \`a partir d'incrustation de signatures de v\'ehicules dans des fonds de sc\`ene, \'eventuellement avec des occultants. Nous d\'eveloppons un principe permettant de g\'en\'erer des signatures de cibles dans diff\'erentes configurations thermiques. Cela nous permet de g\'en\'erer facilement des bases de donn\'ees pour l'\'evaluation d'algorithmes de d\'etection, reconnaissance et identification.
}

\englishabstract{In this communication, we propose a method which permits to simulate images of targets in infrared imagery by superimposition of vehicle signatures in background, eventually with occultants. We develop a principle which authorizes us to generate different thermal configurations of target signatures. This method enables us to easily generate huge datasets for ATR algorithms performance evaluation.
}

\begin{document}

\maketitle

\section{Introduction}
L'\'evaluation et l'optimisation des param\^etres d'un algorithme de d\'etection, reconnaissance, identification (DRI) de cibles en imagerie infrarouge (IR) est fondamentalement d\'e\-pendante de la qualit\'e et de la disponibilit\'e des bases de donn\'ees images utilisables.

L'acquisition de ce type de bases de donn\'ees pr\'esente un co\^ut relativement \'elev\'e et implique d'y consacrer un temps important. Une solution tient en l'utilisation de simulateurs de sc\`enes, mais ceux-ci restent co\^uteux en temps de calcul et il est surtout difficile de choisir les diff\'erents param\^etres afin de balayer de mani\`ere exhaustive un maximum de sc\'enarios op\'erationnels.

Nous proposons de g\'en\'erer de fa\c{c}on hybride ces bases de donn\'ees par incrustation de cibles et occultants sur un fond de sc\`ene sous contrainte de m\'etriques de qualit\'e image. Les param\`etres d'entr\'ee de ces contraintes \'etant les plus efficaces pour d\'ecrire des sc\'enarios op\'erationnels r\'ealistes. De plus, un aspect important de l'imagerie IR tient en la variabilit\'e intrins\`eque de la signature d'une cible. Nous proposons dans cet article une m\'ethode originale permettant de tenir compte de cette variabilit\'e lors de la g\'en\'eration des images. Pour cela, nous utilisons des images r\'eelles de chaque v\'ehicule acquises dans leur mode de fonctionnement \og extr\^emes\fg: cible \`a temp\'erature ambiante et cible avec tous ses \'el\'ements potentiellement en fonction au maximum de leur temp\'erature. Ces signatures sont ensuite plaqu\'ees sur un mod\`ele 3D du v\'ehicule segment\'e en sous-\'el\'ements de signature ayant un comportement thermique jug\'e homog\`ene et ind\'ependant. Il est ainsi possible de param\^etrer s\'electivement le niveau thermique de ces sous-parties, pour construire des variantes de la signature d'une m\^eme cible. Nous utilisons une projection 2D de ce mod\`ele sous l'angle de vue d\'esir\'e pour enfin l'incruster sous contrainte et y appliquer l'effet capteur souhait\'e.

Nous commen\c{c}ons par rappeler le principe de la g\'en\'eration de sc\`ene hybride propos\'ee dans de pr\'ec\'edents travaux \cite{landeau} dans la section \ref{sec:hyb}. Dans la section \ref{sec:var}, nous d\'ecrirons en d\'etail la m\'ethode permettant de cr\'eer les signatures de v\'ehicules en tenant compte de la variabilit\'e thermique intrins\`eque de la cible. En section \ref{sec:res}, nous exposerons divers r\'esultats obtenus par la m\'ethode propos\'ee et enfin nous conclurons en section \ref{sec:con}.

\section{G\'en\'eration hybride de sc\`enes}\label{sec:hyb}
\begin{figure}[t]
\begin{center}
\includegraphics[width=0.45\textwidth]{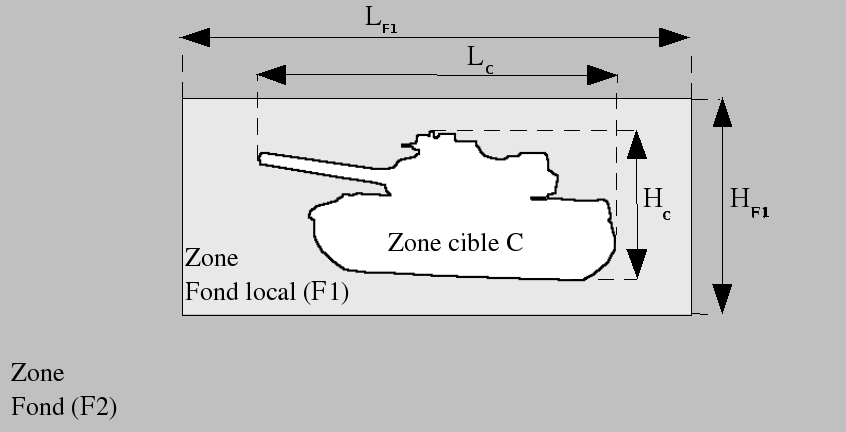}
\end{center}
\caption{D\'efinition des zones d'int\'er\^et pour l'incrustation d'une cible dans un fond.}
\label{fig:zones}
\end{figure}

\begin{figure}[t]
\begin{center}
\includegraphics[width=0.45\textwidth]{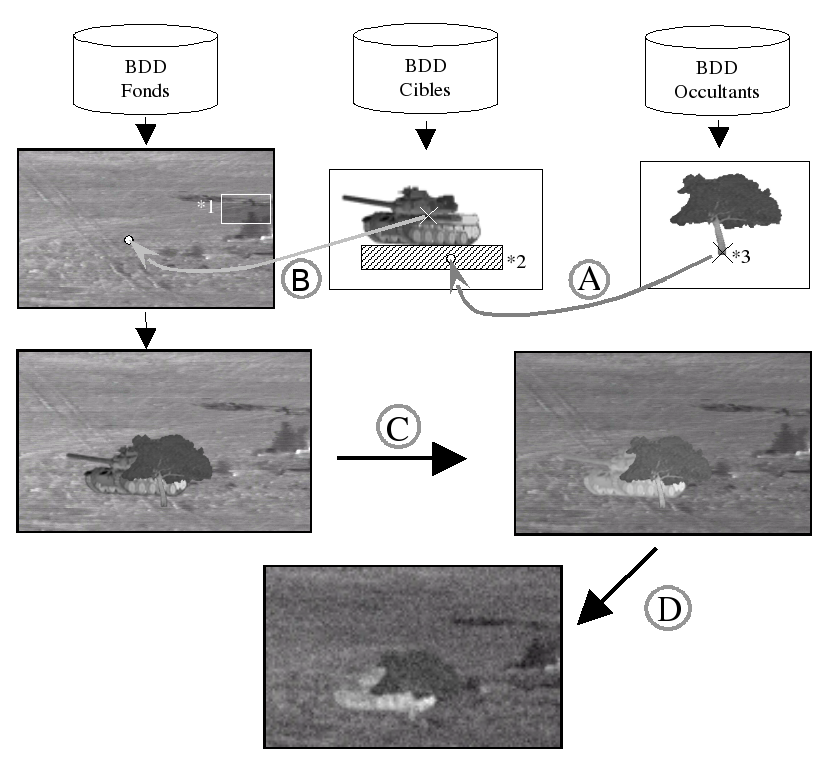}
\end{center}
\caption{Principe de la g\'en\'eration hybride de sc\`ene.}
\label{fig:gene}
\end{figure}

Dans cette section, nous rappelons le principe de g\'en\'eration hybride de sc\`ene propos\'ee dans \cite{landeau}. Cette g\'en\'eration est dite "hybride" car elle consiste \`a incruster une image de cible r\'eelle dans une image de fond, \'eventuellement en positionnant des occultants (arbres, rochers, $\ldots$). L'int\'er\^et de la m\'ethode est qu'il est possible de contr\^oler la qualit\'e de l'image de sortie \`a l'aide de diverses m\'etriques \cite{driggers,nvesd}. Les m\'etriques utilis\'ees sont: le contraste local ($RSS$), la quantit\'e de d\'etectabilit\'e $Q_D$, le rapport signal/fouillis $RSC$, le taux d'occultation $R_x$ et le contraste interne de la cible $K$. Ces quantit\'es sont d\'efinies par

\begin{align}
RSS&=\frac{1}{\nu_k}\sqrt{(\mu_{C}-\mu_{F_1})^2+\sigma_{C}^2} \\
Q_D&=RSS.S_{C} \\
RSC&=\frac{\nu_kRSS}{\sigma_F} \\
R_x&=\frac{S_{\text{zone occult�e}}}{S_{\text{cible totale}}} \\
K&=\frac{\mu_{F_1}-\mu_{C}}{\nu_kRSS}=\frac{\Delta\mu}{\nu_kRSS}
\end{align}

o\`u $C$ d\'esigne la cible, $F_1$ le fond local autour de $C$ et $F_2$ le reste du fond (nous consid\'erons le fond global $F=F_1\cup F_2$), voir la figure \ref{fig:zones}. Alors les quantit\'es $S_x$, $\mu_x$, $\sigma_x$ sont respectivement la surface, la moyenne et l'\'ecart-type de la zone $x$ o\`u $x$ vaut $C$, $F_1$ ou $F_2$. Le coefficient $\nu_k$ est le coefficient permettant de faire la conversion entre niveaux de gris et temp\'erature en kelvin. Le choix de ces param\^etres permet ensuite de calculer des gains et offsets \`a appliquer sur les niveaux de gris de la cible et du fond afin d'obtenir l'image r\'esultante. Enfin, un effet capteur (FTM + \'echantillonnage + bruit) est appliqu\'e. Le processus de g\'en\'eration hybride de sc\`enes est r\'esum\'e sur la figure \ref{fig:gene}. Nous commen\c{c}ons par le positionnement de l'occultant \textcircled{A}, puis le positionnement de la cible \textcircled{B}. Nous appliquons les gains et offsets aux histogrammes de chaque r\'egion \textcircled{C}, pour finir en ajoutant l'effet capteur \textcircled{D}. De plus amples d\'etails ainsi que les expressions des gains et offsets \`a appliquer sont disponibles dans \cite{landeau}. Ce principe de g\'en\'eration de sc\`ene a \'et\'e utilis\'e pour l'\'evaluation des algorithmes de DRI du projet CALADIOM. Toutefois, un aspect n'appara\^it pas dans cet algorithme: la variabilit\'e thermique intrins\`eque des cibles. Nous proposons dans la section suivante une m\'ethodologie pour ajouter cet aspect \`a la g\'en\'eration hybride de sc\`ene.

\section{Variabilit\'e thermique intrins\`eque des cibles}\label{sec:var}
Dans cette section, nous proposons une m\'ethodologie permettant de prendre en compte la variabilit\'e thermique intrins\`eque d'une cible en imagerie IR. En effet, une m\^eme cible peut avoir des apparences thermiques tr\`es diff\'erentes suivant son activit\'e. Par exemple un v\'ehicule \`a l'arr\^et depuis un certain temps n'aura pas ses roues aussi chaudes qu'un v\'ehicule qui vient juste de s'arr\^eter. Or les algorithmes actuels de reconnaissance de v\'ehicule
utilisent la notion d'apprentissage. Il est bien entendu qu'un algorithme de reconnaissance aura plus de difficult\'es \`a donner un r\'esultat pertinent si il n'a jamais appris la cible qui se pr\'esente devant lui (m\^eme s'il l'a d\'ej\`a vu sous un autre aspect thermique).\\

\begin{figure}[t]
\begin{center}
\includegraphics[width=0.35\textwidth]{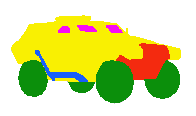}
\end{center}
\caption{Cartographie du comportement thermique type d'un v\'ehicule.}
\label{fig:decoup}
\end{figure}

Cette variabilit\'e est intrins\`eque au fonctionnement du v\'ehicule, cela signifie qu'il suffit de modifier la signature thermique du v\'ehicule. Comme il n'est pas tr\`es r\'ealiste d'un point de vue pratique d'utiliser des mod\`eles physiques de simulation thermique d'un v\'ehicule, nous proposons de cr\'eer des signatures interm\'ediaires \`a partir de signatures \`a temp\'erature ambiante ($TA$) et temp\'erature de fonctionnement ($TF$). Pour cela, nous disposons des mod\`eles 3D de v\'ehi\-cu\-les sur lesquels nous plaquons des textures infrarouges aux niveaux $TA$ et $TF$. Nous proposons d'\'etablir un d\'ecoupage du v\'ehicule en zones ayant des comportements thermiques d\'ependant du mode de fonctionnement du v\'ehi\-cule. Les zones pertinentes retenues sont le moteur, la caisse, l'\'echappement, les vitres et les roues/chenilles (voir la figure \ref{fig:decoup}).

L'\'etat thermique interm\'ediaire $TI$ d'une r\'egion $R$, re\-pr\'e\-sentatif de la variabilit\'e souhait\'ee, est g\'en\'er\'ee en mixant les \'etats $TA$ et $TF$ suivant la relation
\begin{equation}
TI_R=(1-\lambda_R)TA_R+\lambda TF_R,
\end{equation}
o\`u le coefficient $\lambda\in [0;1]$ repr\'esente le niveau de variabilit\'e. Nous d\'efinissons trois zones distinctes de comportement:
\begin{enumerate}
\item en temp\'erature ambiante: $\lambda\in[0;0.1]$,
\item en temp\'erature interm\'ediaire: $\lambda\in]0.1;0.9[$,
\item en temp\'erature de fonctionnement: $\lambda\in[0.9;1]$.
\end{enumerate}

\begin{figure}[t]
\begin{center}
\includegraphics[width=0.45\textwidth]{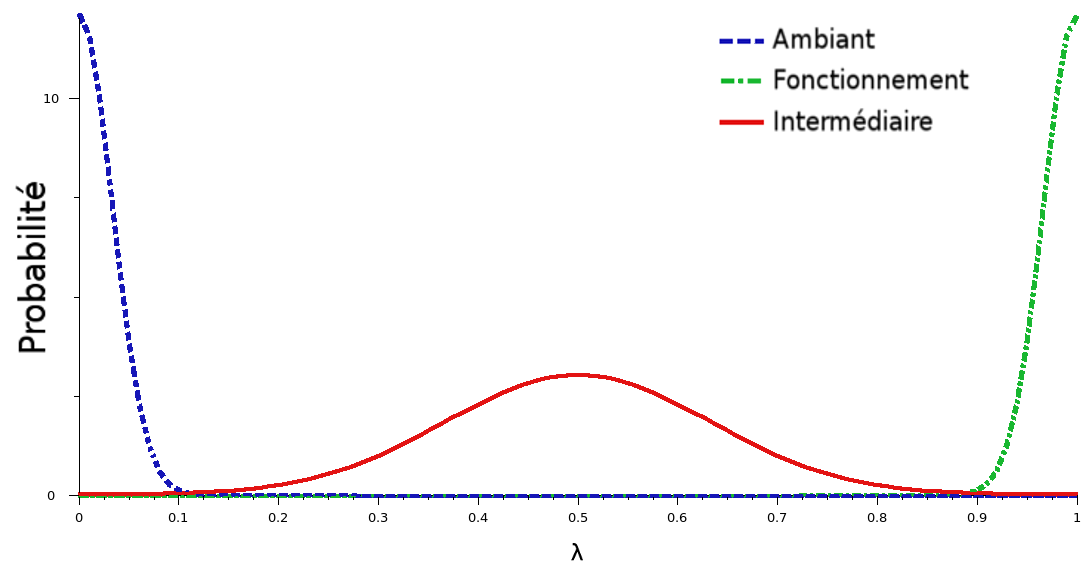}
\end{center}
\caption{Lois de probabilit\'e de $\lambda$ pour chaque mode de fonctionnement.}
\label{fig:gauss}
\end{figure}

Le choix final de la valeur de $\lambda$ se faisant par un tirage al\'eatoire suivant des lois gaussiennes (ou demi-gaussienne aux extr\'emit\'es, voir la figure \ref{fig:gauss}). On choisit la variance de chaque gaussienne de telle sorte que $99\%$ de sa surface soit contenue dans l'intervalle consid\'er\'e. Ceci est \'equivalent \`a dire que $3\sigma_{TA}=3\sigma_{TF}=0.1$ et $3\sigma_{TI}=0.4$, ce qui nous donne respectivement $\sigma_{TA}=\sigma_{TF}=0.33$ et $\sigma_{TI}=0.133$. Les lois sont donc donn\'ees par les expressions (\ref{eq:pta}), (\ref{eq:ptf}) et (\ref{eq:pti}) (pour les valeurs de $\lambda$ prises dans les intervalles d\'efinis pr\'ec\'edemment).

\begin{align}
P_{TA}(\lambda)&=\frac{1}{\sqrt{2\pi\sigma_{TA}^2}}\exp(-\frac{\lambda^2}{2\sigma_{TA}^2}) \label{eq:pta} \\
P_{TF}(\lambda)&=\frac{1}{\sqrt{2\pi\sigma_{TF}^2}}\exp(-\frac{(1-\lambda)^2}{2\sigma_{TF}^2}) \label{eq:ptf} \\
P_{TI}(\lambda)&=\frac{1}{\sqrt{2\pi\sigma_{TI}^2}}\exp(-\frac{(\lambda-0.5)^2}{2\sigma_{TI}^2}) \label{eq:pti}
\end{align}

En se fixant diff\'erentes configurations de fonctionnement (par exemple pour un v\'ehicule \`a l'arr\^et, moteur allum\'e, on aurait caisse, vitres et roues \`a temp\'erature ambiante et moteur et \'echappement \`a temp\'erature de fonctionnement), nous pouvons g\'en\'erer les textures \`a plaquer sur le mod\`ele 3D. Il reste ensuite \`a g\'en\'erer des images du v\'ehicule pour diff\'erents points de vue et ainsi \'etoffer la base de donn\'ees de signatures de v\'ehicules en entr\'ee de l'algorithme de g\'en\'eration hybride de sc\`ene.

\section{R\'esultats}\label{sec:res}
Dans cette section, nous pr\'esentons quelques r\'esultats ob\-tenus gr\^ace \`a la m\'ethode d\'ecrite pr\'ec\'edemment. 

\begin{figure}[t]
\begin{tabular}{cc}
\includegraphics[width=0.22\textwidth]{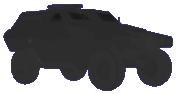} & \includegraphics[width=0.22\textwidth]{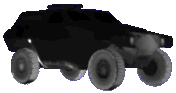} \\
\includegraphics[width=0.22\textwidth]{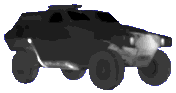} & \includegraphics[width=0.22\textwidth]{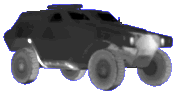}
\end{tabular}
\caption{Exemple de configurations thermiques g\'en\'er\'ees par la m\'ethode.}
\label{fig:sign}
\end{figure}

En premier lieu, la figure \ref{fig:sign} pr\'esente diff\'erentes configurations thermiques d'un m\^eme v\'ehicule suivant le m\^eme point de vue. Nous voyons qu'il nous est donc possible d'avoir des signatures r\'ealistes correspondant \`a certains domaines d'emploi (v\'ehi\-cule totalement \`a l'arr\^et, v\'ehicule mobile, v\'ehicule en poste, $\ldots$).
Nous pouvons donc g\'en\'erer une quelconque vue d'un v\'ehicule par cette m\'ethode.

\begin{figure}[t]
\begin{center}
\includegraphics[width=0.35\textwidth]{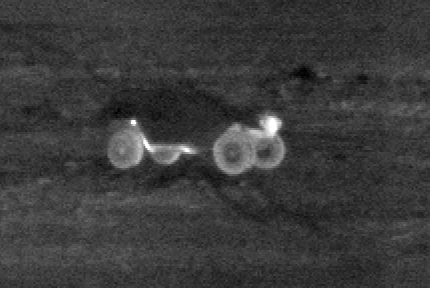} 
\includegraphics[width=0.35\textwidth]{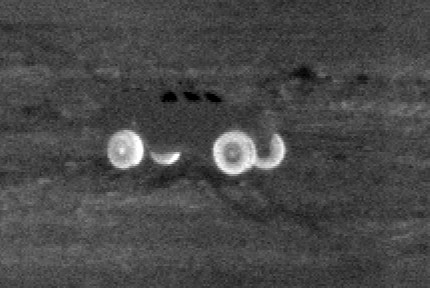} 
\includegraphics[width=0.35\textwidth]{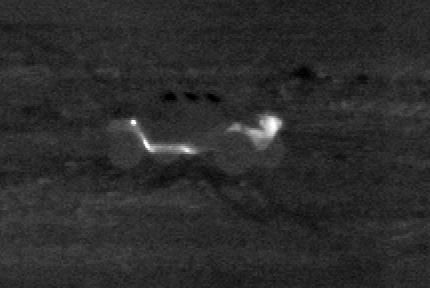} 
\caption{Exemple d'une m\^eme sc\`ene avec diff\'erentes configurations thermiques d'un v\'ehicule.}
\label{fig:incr}
\end{center}
\end{figure}

Dans un deuxi\`eme temps, ces nouvelles signatures constituent une nouvelle base de donn\'ees pour le g\'en\'erateur hybride de sc\`ene. Nous pouvons alors cr\'eer des sc\`enes en tenant compte de la variabilit\'e thermique intrins\`eque des cibles en incrustant des cibles choisies dans cette nouvelle base de donn\'ee. La figure \ref{fig:incr} pr\'esente une m\^eme sc\`ene, g\'en\'er\'ee avec les m\^eme param\`etres de qualit\'e image, contenant le m\^eme v\'ehicule mais avec des configurations thermiques diff\'erentes. 

\begin{figure}[t]
\begin{center}
\includegraphics[width=0.25\textwidth]{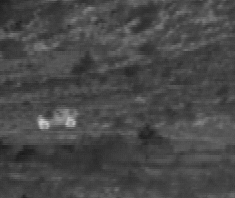} 
\includegraphics[width=0.4\textwidth]{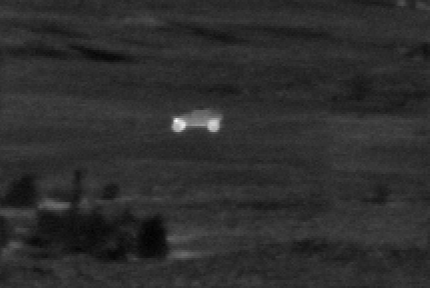} 
\caption{Exemple d'une acquisition r\'eelle (en haut) et d'une image simul\'ee (en bas).}
\label{fig:reel}
\end{center}
\end{figure}

Par ailleurs, la figure \ref{fig:reel} illustre un m\^eme v\'ehicule sur une image r\'eelle acquise lors d'une campagne d'essai (en haut) et sur une image simul\'ee en sortie de la m\'ethode propos\'ee (en bas). Ceci montre le r\'ealisme des simulations obtenues et l'int\'er\^et de la m\'ethode.

\section{Conclusion}\label{sec:con}
Dans cet article, nous pr\'esentons une m\'ethodologie permettant de g\'en\'erer des images de sc\`enes infrarouge r\'ealistes. Le principe consiste \`a incruster des signatures de v\'ehicules sur des fonds de sc\`ene, \'eventuellement en ajou\-tant des occultants, le tout sous contrainte de param\^etres de qualit\'e image. Nous abordons par ailleurs la prise en comp\-te de la variabilit\'e thermique intrins\`eque des cibles. Pour cela, nous proposons d'interpoler des signatures ther\-miques \`a partir de la connaissance des signatures \`a temp\'erature ambiante et de fonctionnement. La signature r\'esultante est alors plaqu\'ee sur un mod\`ele 3D permettant de g\'en\'erer des prises de vues suivant l'angle de vision voulu.

Cette m\'ethode permet de construire des bases de donn\'ees d'images pour la mise au point et l'\'evaluation d'algo\-rithmes de d\'etection, reconnaissance et identification de cibles. 

L'avantage de la m\'ethode propos\'ee est qu'il devient ais\'e de construire des bases de donn\'ees exhaustives tout en ayant une \'evaluation quantitative des performances de l'algorithme. A l'heure actuelle, nous avons mis \`a disposition des \'equipes de d\'eveloppement d'algorithmes une base de donn\'ees de plus de dix mille images de cibles avec variabilit\'e thermique et leurs v\'erit\'es terrain associ\'ees.

Nous \'etendons la m\'ethode \`a la g\'en\'eration des s\'equen\-ces d'images avec cible mobile en vue de l'\'evaluation d'algo\-rithmes de poursuite de cibles. Le principe \'etant de fournir une trajectoire d\'efinie a priori. L'algorithme se charge alors de calculer l'angle de vue de la cible par rapport \`a la position du capteur, lui permettant d'aller s\'electionner dans la base de donn\'ees de cibles la signature ad\'equate \`a incruster.


\begin{thebibliography}{99}
\bibitem{landeau} S.Landeau, T.Dagobert, \og{}Image database generation using image metric constraints: an application within the CALADIOM project\fg{}, SPIE Security and Defense, Orlando (US-Florida), 2006.
\bibitem{driggers} R.Driggers, P.Cox, T.Edwards, \og{}Introduction to infrared and electro-optical systems\fg{}.
\bibitem{nvesd} Vollmerhausen, E.Jacobs, Hixon, Friedman, \og{}NVTherm IP, the targetting task performance (TTP) metric\fg{}, Technical report AMSEL-NV-TR-230.
\end{thebibliography}
\end{document}